\documentclass[letterpaper, 10 pt, conference]{ieeeconf}  

\IEEEoverridecommandlockouts                              

\usepackage{graphics} 
\usepackage{epsfig} 
\usepackage{mathptmx} 
\usepackage{times} 
\usepackage{amsmath} 
\usepackage{amssymb}  
\usepackage{amsfonts}
\usepackage{mathtools}
\usepackage{commath}
\usepackage{graphicx}
\usepackage{algorithm,algorithmic}
\usepackage[bookmarks=false]{hyperref}
\usepackage[font=small]{caption}
\usepackage{subcaption}
\usepackage{threeparttable}

\usepackage[noadjust]{cite}

\def\equationautorefname~#1\null{(#1)\null}
\def\sectionautorefname~#1\null{section #1\null}
\def\subsectionautorefname~#1\null{section #1\null}
\def\algorithmautorefname~#1\null{algorithm #1\null}
\makeatletter
\def\ALC@uniqueautorefname~#1\null{line #1\null}
\makeatother

\usepackage{kotex}

\usepackage{xcolor}

\usepackage{tikz}

\newcommand\copyrighttext{%
  \footnotesize \textcopyright 2021 IEEE.  Personal use of this material is permitted.  Permission from IEEE must be obtained for all other uses, in any current or future media, including reprinting/republishing this material for advertising or promotional purposes, creating new collective works, for resale or redistribution to servers or lists, or reuse of any copyrighted component of this work in other works.}
\newcommand\copyrightnotice{%
\begin{tikzpicture}[remember picture,overlay]
\node[anchor=south,yshift=10pt] at (current page.south) {\fbox{\parbox{\dimexpr\textwidth-\fboxsep-\fboxrule\relax}{\copyrighttext}}};
\end{tikzpicture}%
}

\title{\LARGE \bf
Robust and Recursively Feasible Real-Time Trajectory Planning\\ in Unknown Environments
}

\author{Inkyu Jang, Dongjae Lee, Seungjae Lee, and H. Jin Kim
\thanks{This research was supported by Unmanned Vehicles Core Technology Research and Development Program through the National Research Foundation of Korea (NRF) and Unmanned Vehicle Advanced Research Center (UVARC) funded by the Ministry of Science and ICT, the Republic of Korea (NRF-2020M3C1C1A01086411).}
\thanks{The authors are with the Department of Aerospace Engineering, and Automation and Systems Research Institute (ASRI), Seoul National University, Seoul, Korea.
        {\tt\small \{leplusbon, ehdwo713, ysz0301, hjinkim\}@snu.ac.kr}}%
}

\begin{document}

\maketitle
\copyrightnotice 
\thispagestyle{empty}
\pagestyle{empty}

\begin{abstract}

Motion planners for mobile robots in unknown environments face the challenge of simultaneously maintaining both robustness against unmodeled uncertainties and persistent feasibility of the trajectory-finding problem. That is, while dealing with uncertainties, a motion planner must update its trajectory, adapting to the newly revealed environment in real-time; failing to do so may involve unsafe circumstances. Many existing planning algorithms guarantee these by maintaining the clearance needed to perform an emergency brake, which is itself a robust and persistently feasible maneuver. However, such maneuvers are not applicable for systems in which braking is impossible or risky, such as fixed-wing aircraft. To that end, we propose a real-time robust planner that recursively guarantees persistent feasibility without any need of braking. The planner ensures robustness against bounded uncertainties and persistent feasibility by constructing a loop of sequentially composed funnels, starting from the receding horizon local trajectory's forward reachable set. We implement the proposed algorithm for a robotic car tracking a speed-fixed reference trajectory. The experiment results show that the proposed algorithm can be run at faster than 16 Hz, while successfully keeping the system away from entering any dead end, to maintain safety and feasibility.

\end{abstract}

\section{Introduction}

Motion planners for autonomous mobile robots in unknown space should tackle two major challenges: being \textit{robust} against unmodeled uncertainties, and persistently maintaining the feasibility of the planning problem itself.

Unmodeled uncertainties include the trajectory tracking error, sensor measurement error, and/or external forces such as wind disturbance.
Their influence is often overcome by employing a trajectory planner which is robust against a prescribed class of uncertainties \cite{hoseong_hj1, fastrack, majumdar2013robust}, e.g., wind disturbance of known maximum speed. Such methodologies provide a tube, within which the system is guaranteed to stay without collision. However, these tube-based methodologies often drive the system into dead ends from which the system cannot escape. Moreover, newly discovered obstacles may invalidate the trajectory. In such cases, failing to find a feasible trajectory update might lead to loss of safety.

Therefore, maintaining \textit{persistent feasibility} is as important.
It is usually ensured through recursion \cite{liniger2015viability, recursive_feasibility_icra2019}: i.e., \textit{recursive feasibility} holds if having a single feasible maneuver provides either persistent safety or the feasibility of finding a next one.
For many systems, braking is a widely-used recursively feasible maneuvering strategy. \cite{majumdar2017funnel} assumes the system's ability to perform a sudden stop, and \cite{faster, kousik2020bridging} explicitly compute the spare space needed for the system to brake.
However, in some systems such as a fixed-wing aircraft, the mentioned emergency brakes are difficult or impossible. Feasibility must therefore be considered more carefully for them.
\begin{figure}
    \centering
    \includegraphics[width=0.9\linewidth]{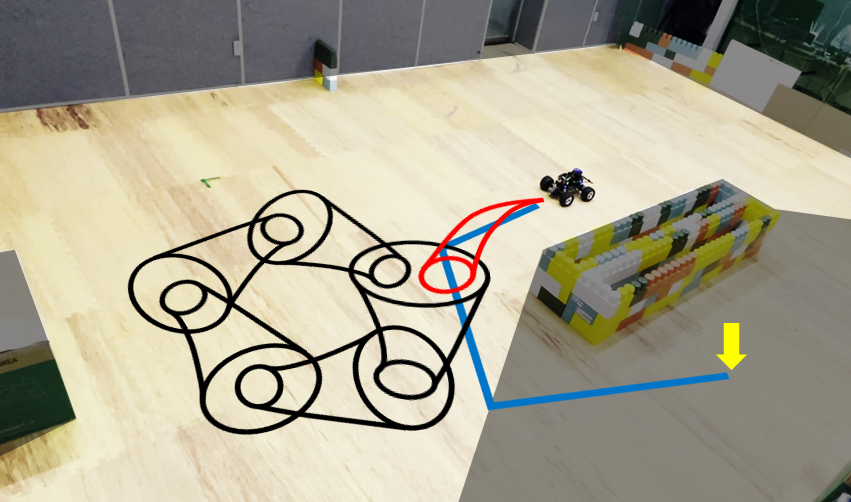}
    \caption{An autonomous ground rover avoids obstacle while trying to reach the goal position (marked using yellow arrow) in unknown space (shadowed region). The global path that leads to the goal is marked blue. The FRS around the local receding horizon trajectory is marked red. Starting from the endpoint of the local trajectory, funnels are cyclically composed within known free area to ensure recursive feasibility of the trajectory planner.}
    \vspace{-0.7cm}
    \label{fig:thumbnail}
\end{figure}

In this paper, we present a real-time trajectory planning algorithm that is both robust to bounded uncertainties and recursively feasible at the same time, without any need of braking capability.
We begin by following the common planning framework, planning a global path to the goal through the free and unknown spaces, which is tracked by a receding horizon local trajectory in the free space.
We guarantee recursive feasibility by attaching a loop of sequentially constructed funnels starting from the local trajectory. Robustness against uncertainties is guaranteed by requiring the forward reachable set (FRS) of the local trajectory to end inside the funnel loop.
Suggested in \cite{majumdar2017funnel}, funnels are basically FRSs that start from a big initial set. If the system is located inside a funnel's initial set, it can be provably driven to its exit by its associated controller. The set of pre-computed funnels, funnel library, can be used as building blocks for safe trajectory. If a funnel's entrance encompasses another funnel's exit, than the two funnels are sequentially composed and act like a single funnel. 
\autoref{fig:thumbnail} briefly shows how the proposed planner runs. At an epoch, the planner finds the global and local trajectories, the FRS of the local trajectory, and a funnel loop that follows the local trajectory. A funnel loop acts like a single funnel whose entrance encompasses its exit, so the system can permanently stay inside the loop without braking.



\subsection{Contributions}

The contributions of this paper can be summarized as follows.
\begin{itemize}
    \item We propose a trajectory planner that utilizes cyclic funnel compositions to guarantee persistent safety and feasibility. 
    \item We present a forward reachability analysis method that simulates the adversarial disturbance sequence in real-time, in order to check whether the system can robustly be driven into the funnel loop or not.
    \item The planner is scalable in execution time. It can be run long without any need of additional memory, since we discard unnecessary funnel loops once another is found.
    \item Our planner provides some resilience to map changes. 
    Since it creates the funnel loop only in the neighborhood of the current configuration, persistent feasibility is still guaranteed as long as map changes only occur outside the existing funnel loop.
    \item The proposed algorithm is validated through an experiment using an actual ground robot running an off-the-shelf mapping software. The proposed planner achieves computation time of less than 60 ms and can be run at real-time. 
\end{itemize}

\subsection{Relevant Work}

The influence caused by uncertainties is often overcome by having a reachability-based robust motion planner. Such planners pre-compute the FRS of the system, with known disturbance bounds. 
The most accurate approximation of the FRS is obtained by solving the Hamilton-Jacobi partial differential equations (PDE) \cite{hjreachability_review, ding2011reachability}, although solving them requires extremely burdensome computation.
For polynomial dynamics, one can achieve faster FRS computation by converting the complex PDE into a convex optimization problem. Sum-of-squares (SOS) programming was used in \cite{majumdar2017funnel}, and to further accelerate computation speed, positive polynomials such as Bernstein bases can be used \cite{bernsteinfunnel}. However, due to the limited expressivity of polynomials, both entail over-conservatism as a trade-off. Some planners circumvent burdensome computations by using pre-calculated asymptotic bounds on which feedback control and disturbance effects are balanced \cite{fastrack, singh2017robust}.
\cite{villanueva2015unified, villanueva2017robust} propose tube-based model predictive control (MPC) approaches that computes the bound online through min-max inequalities, but they carry heavy computational load.

In the perspective of motion planning, \cite{majumdar2017funnel} composes funnels sequentially to build a trajectory in known environments from start to goal. The funnels can be also composed in runtime (not in the planning phase), but it requires the robot to be able to perform a sudden stop in case of an infeasibility. In \cite{hoseong_hj1} and \cite{hoseong_hj2}, Hamilton-Jacobi reachability is directly used. The partial differential equation is translated to a more conservative but simpler form, so that an over-approximated FRS can be computed in real-time along the planned trajectory.


To take persistent feasibility into account, \cite{fraichard2004inevitable} identifies inevitable collision states, which are instantaneously collision-free but will unavoidably lead to collision with a nearby obstacle. Avoiding such states provides safety.
\cite{faster, oleynikova2018safe} have separate local planners to find a trajectory with final stop condition within the known-free space. In \cite{kousik2020bridging}, the braking maneuver is explicitly considered to guarantee persistent feasibility.
\cite{loiter_box_aiaa} addresses the problem for multiple airplanes. Each agent avoids collision and infeasibility by having a virtual box on one side that fits a loiter pattern and letting nothing inside the box.
In \cite{recursive_feasibility_icra2019}, the system expands the exploration tree consisting only of points from which the agent can safely return to the home position. The algorithm can be applied to systems with no brakes, but is vulnerable to even small changes in obstacle configuration near home, because this change may block many returning paths.
In the world of reinforcement learning, in \cite{leave_no_trace}, the agents learn how to reset themselves when they encounter irreversible or hazardous states. Penalizing resets will eventually lead the agent to learn a reversible policy, although this is out of the scope of this paper.



\section{Preliminaries}

We start by considering a nonlinear discrete-time time-invariant mobile robot system with state $x \in \mathbb{R}^n$ and input $u \in \mathbb{U} \subseteq \mathbb{R}^m$ under additive disturbance $w \in \mathbb{W} \subseteq \mathbb{R}^n$. The evolution of the system at any epoch $t$ is governed by
\begin{equation}
    x(t+1) = f(x(t), u(t)) + w(t).
\end{equation}
The state space consists of two components: cyclic and noncyclic coordinates. Let $x = (x_{\mathrm{c}}, x_{\mathrm{nc}}) \in \mathbb{R}^n$, where $x_{\mathrm{c}} \in \mathbb{R}^d$ and $x_{\mathrm{nc}} \in \mathbb{R}^{n-d}$ are the cyclic and noncyclic parts, respectively. The system's governing equation is assumed to be invariant under translation in the cyclic coordinates, i.e., we can write
\begin{equation}
\begin{aligned}
    x(t+1) &= f(x(t), u(t)) + w(t) \\
    &= x(t) + \hat{f}(x_{\mathrm{nc}}(t), u(t)) + w(t),
\end{aligned}
\end{equation}
where $\hat{f}:\mathbb{R}^{n-d}\times \mathbb{R}^m \rightarrow \mathbb{R}^n$ is a function only of noncyclic coordinates and input.
In mobile robot systems, the cyclic coordinates are usually their position in the Euclidean workspace. The cyclic component can be obtained using the projection map $\Pi:\mathbb{R}^n\rightarrow \mathbb{R}^d$:
\begin{equation}
    x_{\mathrm{c}} = \Pi (x).
\end{equation}
The remaining parts of this paper assume that the cyclic coordinates of the system is equal to the robot's position, hence collision checking can be done by just examining the cyclic coordinates. 






A funnel $\mathcal{F}=(I, E, X)$ is a tuple consisting of three sets: entrance $I \subseteq \mathbb{R}^n$ and exit $X \subseteq \mathbb{R}^n$ in the state space, encompassing shape $E \subseteq \mathbb{R}^d$ in the Euclidean workspace. Mathematically, if $x(t_0) \in I$, then there exists an epoch $t_1>t_0$ and $u(t, x(t), \cdots) \in \mathbb{U} \;\;, \forall t \in [t_0, t_1]$, such that $x(t_1) \in X$ and $\Pi (x(t)) \in E, \;\; \forall t \in [t_0, t_1]$. In this paper, we consider funnels that satisfy the following:
\begin{itemize}
    \item The entrance $I$ is of the form $I = \{(p, x_{\mathrm{nc}}) \in \mathbb{R}^n \;|\; p \in \Pi(I), x_{\mathrm{nc}} \in I_{\mathrm{nc}} \subseteq \mathbb{R}^{n-d} \}$, where $\Pi(I) = \{p \in \mathbb{R}^d\;|\; A_I (p-p_I) \leq b_I\}$. $A_I$ and $b_I$ are matrix and column vector with appropriate sizes, respectively. That is, the constraints for cyclic and noncyclic coordinates are decoupled, and the cyclic coordinate part is a polytope.
    $p_I \in \mathbb{R}^d$ is the \textit{center} of $I$, which is the starting point of the nominal trajectory that traverses the funnel.
    \item The exit $X$ is of the form $X=\{(p, x_{\mathrm{nc}}) \in \mathbb{X} \;|\; \norm{p - p_X}_2 \leq r_X, x_{\mathrm{nc}} \in X_{\mathrm{nc}} \subseteq \mathbb{R}^{n-d}\}$, where $p_X$ is the (cyclic) coordinate of the funnel exit, $r_X > 0$ is the exit radius on the cyclic coordinate.
    \item The encompassing shape $E$ is a polytope, i.e., $E=\{p \in \mathbb{R}^d\;|\; A_E p \leq b_E\}$, where $A_E$ is a matrix and $b_E$ is a column vector, both with appropriate sizes. 
\end{itemize}
Additionally, for the sake of simple description, we normalize each row of $A_I$ and $A_E$ to be a unit vector.

We say that two funnels $\mathcal{F}_i = (I_i, X_i, E_i)$ and $\mathcal{F}_j = (I_j, X_j, E_j)$ are sequentially composed if the entrance of $\mathcal{F}_j$ completely encompasses $X_i$, i.e., 
\begin{equation} \label{eq:x_in_i}
    X_i \subseteq I_j,
\end{equation}
so that starting in $I_i$ guarantees the existence of a trajectory that leads to $X_j$.
We denote this using the $\rhd$ symbol, i.e., $\mathcal{F}_i \rhd \mathcal{F}_j$. 
If $\mathcal{F}_1 \rhd \cdots \rhd \mathcal{F}_{n_F}$ and $\mathcal{F}_{n_F} \rhd \mathcal{F}_1$, the funnels form a loop, which the system can persistently stay within.

\section{Algorithm Overview}

\begin{figure}[t]
    \centering
    \includegraphics[width=0.9\linewidth]{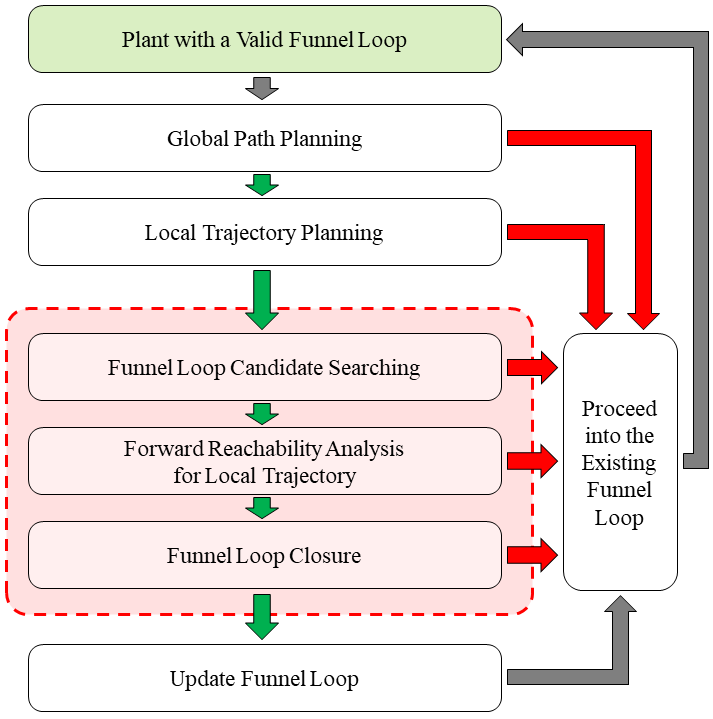}
    \caption{A schematic explanation of the proposed planning algorithm. Green and red arrows denote that the corresponding step has succeeded or failed, respectively. The funnel loop finding part is marked using a red box.}
    \vspace{-0.3cm}
    \label{fig:planningalgorithm}
\end{figure}

\begin{figure}[t]
    \centering
    \includegraphics[width=0.43\textwidth]{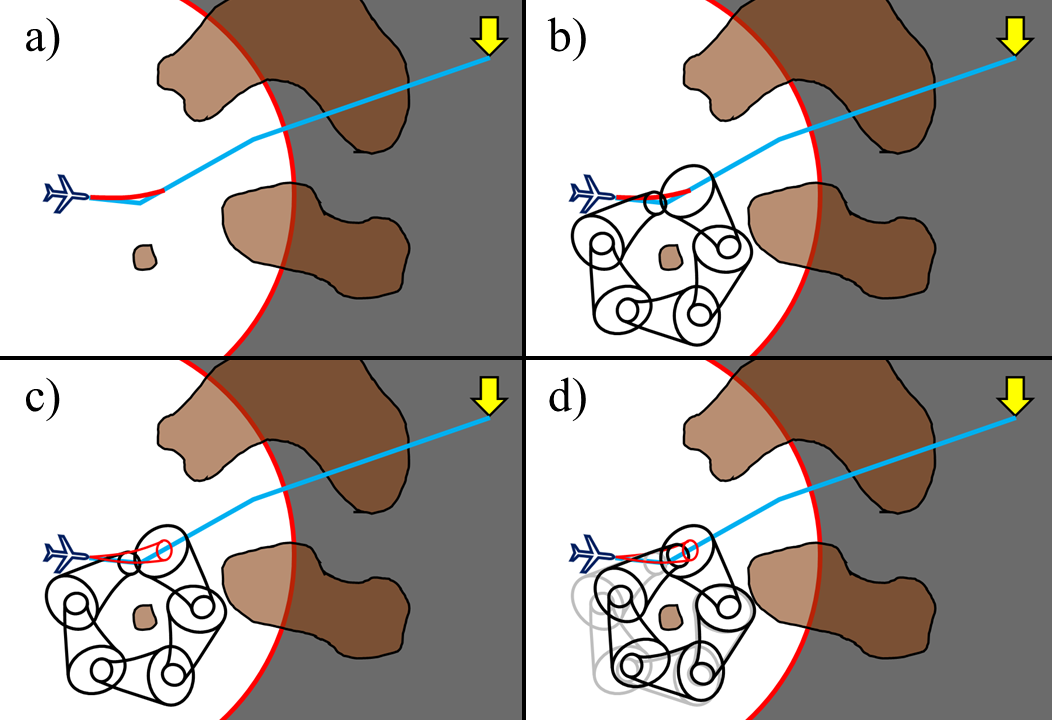}
    \caption{A graphical description of the funnel planning algorithm (the red-boxed part of \autoref{fig:planningalgorithm}). The system is drawn as an airplane.
    The obstacle areas are colored brown, and only the bright regions are known to the system. The goal position is marked with yellow arrow. a) A global path (blue curve) and local trajectory (red short curve) are found using the global and local planners. b) A funnel loop candidate is found by growing a tree of funnels. c) Forward reachability analysis ensures that the local trajectory is free from collision and safely leads the system into the first funnel's entrance. d) The funnel positions are adjusted to complete a closed loop.}
    \vspace{-0.3cm}
    \label{fig:algorithm}
\end{figure}

Our algorithm adds the following three steps to the global-local trajectory planning algorithm mentioned in the introduction: funnel loop candidate searching, forward reachability analysis, and funnel loop closure.
In the funnel loop candidate searching step, we find a sequence of funnels which are expected to be adjustable in the cyclic coordinates to form a valid loop. The forward reachability analysis ensures that the planned local trajectory robustly drives the system into the entrance of the first funnel, despite the effect of disturbances. In the last step, the funnel positions are adjusted so that the loop is closed while maintaining collision avoidance and robustness of the local planner.
As a result, we always have a \textit{funnel loop trajectory}, which consists of a collision-free funnel loop and a robust local trajectory that drives the system into the loop. The system can stay inside the funnel loop as long as needed, before it finds a next valid funnel loop trajectory.

A schematic explanation of this procedure is given in \autoref{fig:planningalgorithm}. The proposed three steps are marked using a red box. \autoref{fig:algorithm} provides a graphical explanation of the algorithm. In the next section, we elucidate each step of the proposed algorithm.

\section{Robust Funnel Loop Planning}

For an ordered set of $n_F$ funnels $(\mathcal{F}_1 = (I_1, X_1, E_1)$, $\cdots$, $\mathcal{F}_{n_F} = (I_{n_F}, X_{n_F}, E_{n_F}))$ to form a valid loop, \autoref{eq:x_in_i} should hold for all neighboring index pairs $(i, j) \in \mathbb{I} = \{(1, 2), \cdots, (n_F - 1, n_F), (n_F, 1)\}$. That is, to elaborate, 
\begin{equation} \label{eq:cycliccondition_nc}
    X_{\mathrm{nc}\cdot i} \subseteq I_{\mathrm{nc}\cdot j} \quad \forall (i, j) \in \mathbb{I}
\end{equation}
and
\begin{equation} \label{eq:cycliccondition_c}
    A_{I_j} (p_{X_i} - p_{I_j}) \leq b_{I_j} - r_{X_i} \quad  \forall (i, j) \in \mathbb{I}.
\end{equation}
The goal of this section is to find $n_F$ funnels, each of them being a translated funnel from the funnel library $\mathfrak{F} = \{\mathcal{F}_{L \cdot 1}, \cdots, \mathcal{F}_{L \cdot n_{\mathfrak{F}}}\}$, such that the funnels satisfy \autoref{eq:cycliccondition_nc} and \autoref{eq:cycliccondition_c}.
We tackle this through a three-step process: in step 1, we find the funnel sequence that satisfies \autoref{eq:cycliccondition_nc} completely and \autoref{eq:cycliccondition_c} roughly using a graph search algorithm (\autoref{sec:astar}); in step 2, we confirm that the local trajectory robustly leads to the entrance of the funnel trajectory candidate of step 1 (\autoref{sec:frs});
and in the last step, we adjust the positions of the funnel path candidate to completely satisfy \autoref{eq:cycliccondition_c} (\autoref{sec:adjustment}). This is possible owing to the funnel property that the constraints in the cyclic and noncyclic coordinates are decoupled.

\subsection{Funnel Loop Candidate Searching} \label{sec:astar}

We find a cyclic funnel path candidate by constructing and traversing a tree of properly connected funnels.
First, the tree is initialized with the reachable set of the local trajectory as the root node.
Given a node $X = X_{\mathrm{c}}\times X_{\mathrm{nc}} \subseteq \mathbb{R}^n$ where $X_{\mathrm{c}} = \{p \;|\; \norm{p-p_X}_2 \leq r_X \} \subseteq \mathbb{R}^d$, the funnels that satisfy \autoref{eq:cycliccondition_nc} are chosen and are translated such that the cyclic coordinates of the entrance and $p_X$ are matched. If a translated funnel is collision-free, its exit becomes the child node.
We continue this process until we find a node that is sufficiently close to the root node in the cyclic coordinates, and is completely encompassed by the first funnel's entrance in the noncyclic coordinates. The criterion used to determine whether a node is \textit{sufficiently close} may vary depending on the system or the environment, and can be heuristically selected to maximize the success rate of finding a valid funnel path.
Any off-the-shelf tree searching algorithm can be used in this step. 

\subsection{Forward Reachability Analysis} \label{sec:frs}

After finding a funnel cycle candidate, the validity of the local planning results is checked.
The checking procedure consists of two steps: collision avoidance, and funnel composability.

\begin{figure}
    \centering
    \includegraphics[width=0.4\textwidth]{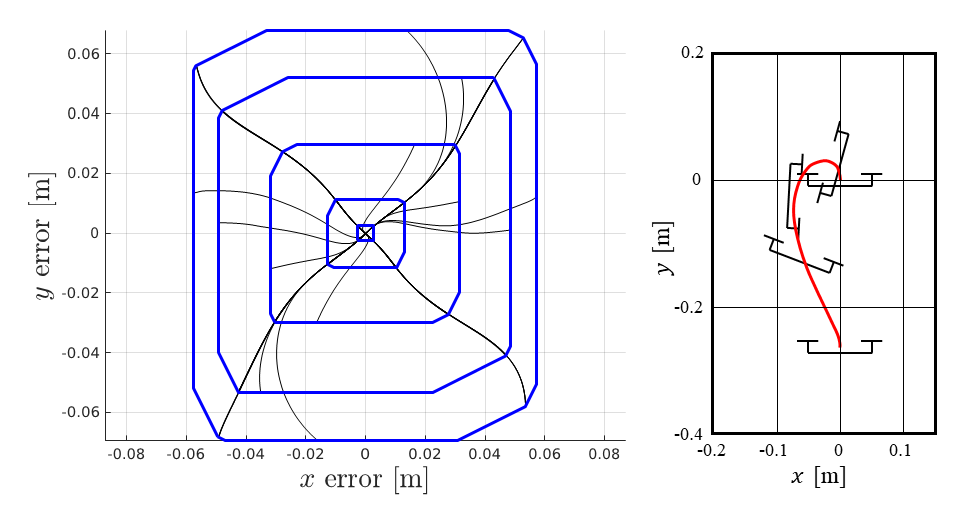}
    \caption{The FRS of a planar quadrotor system with state $(x, y, \theta, \dot{x}, \dot{y}, \dot{\theta}) \in \mathbb{R}^6$ and input $(f, \tau) \in \mathbb{R}^2$ in a flipping maneuver shown in the right, where $x$, $y$, and $\theta$ are the quadrotor's horizontal displacement, altitude, and pitch angle, respectively. The input consists of thrust $f$ and pitching torque $\tau$. The blue bounds represent the estimated FRS, and black curves are the error trajectories caused by adversarial disturbance.}
    \vspace{-0.4cm}
    \label{fig:frs}
\end{figure}

Assume that we are given a memoryless feedback controller $u(t) = k(t; x(t), x_r(t))$ and the local trajectory $x_r(t)\in\mathbb{R}^n$ for $t \in \{0, \cdots, T\}$. 
Then, we have the time-varying closed-loop error dynamics 
\begin{equation} \label{eq:forwardpass}
\begin{aligned}
    e(t+1) {}={}& f(x_r(t) + e(t), \;k(t; x_r(t) + e(t), x_r(t))) \\
    & {} - x_r(t) + w(t) \\
    {}={}& h(t; e(t)) + w(t)
\end{aligned}
\end{equation}
where the error $e(t)\in\mathbb{R}^n$ is defined as $e(t) = x(t)-x_r(t)$, and $h$ is the function that describes the closed-loop dynamics.
We assume that the disturbance $w(t)$ is bounded by $w(t) \in \mathbb{W}=\{w \in \mathbb{R}^n \;|\; A_w w \leq b_w\}$. 
The goal of this step is to find the sequence of disturbance $w(t)$ for $t \in \{0, \cdots, \tau - 1\}$ for an epoch $\tau \in \{1, \cdots, T\}$, which drives the system in the way that maximizes a given objective function $a: \mathbb{R}^n \rightarrow \mathbb{R}$. That is, we solve the following optimal control problem:
\begin{equation} \label{eq:ddp}
\begin{aligned}
    \max. & \quad a(e(\tau)) \\
    \mathrm{s.t.} & \quad e(t + 1) = h(t; e(t)) + w(t) & \forall t \in \{0, \cdots, \tau - 1\} \\
    & \quad e(0) = x_0 - x_r(0) \\
    & \quad w(t) \in \mathbb{W} & \forall t \in \{0, \cdots, \tau - 1\}.
\end{aligned}
\end{equation}
To solve this, we use a modified version of the differential dynamic programming (DDP) algorithm that only propagates the first-order gradients, which is an extension of \cite{iccasfrs} to nonlinear systems. 

First, the algorithm initializes with the disturbance sequence $w(t)$. The error sequence $e(t)$ is determined by simulating \autoref{eq:forwardpass} forwards.
Let $g_t$ denote the gradient of the objective function with respect to $w(t)$, on the forward-passed trajectory. It is straightforward to find out that 
\begin{equation}
    g_{\tau -1} = \left(\frac{\partial a}{\partial e} (e(\tau))\right)^\top.
\end{equation}
We can now backward-pass $g_t$ using
\begin{equation}
    g_{t - 1} = \left(\frac{\partial e(t)}{\partial e(t - 1)}\right)^\top g_{t} = \left(\frac{\partial h(t-1; \cdot)}{\partial e} (e(t - 1))\right)^\top g_{t},
\end{equation}
which can be derived using the chain rule. The disturbance sequence is updated using the steepest ascent by solving the following quadratic programming (QP) problem:
\begin{equation} \label{eq:backwardpass}
    w(t)' = \underset{w \in \mathbb{W}}{\mathrm{argmax.}} \; g_t^\top w + \alpha \norm{w - w(t)}_2^2,
\end{equation}
where $w(t)'$ is the new value for $w(t)$. The weight $\alpha \geq 0$ penalizes drastic changes in $w(t)$ and prevents the problem from falling into a local optimum. We found that in many cases, setting $\alpha = 0$ is sufficient: in such cases, the problem becomes a linear programming (LP) problem. Since the domain $\mathbb{W}$ is the same in every update, we can pre-compute the vertices and their connectivity in the offline phase in order to enhance the computation speed.
Unlike DDP methods used in trajectory planning or MPC, the backward pass can be calculated in parallel, since \autoref{eq:backwardpass} does not reuse the previous calculation results. \autoref{fig:frs} shows an example FRS calculated using the proposed method for linear objective functions.

Now, we denote the entrance of the first funnel by $I_1 = \{(p, x_{\mathrm{nc}}) \;|\; A_{I_1} p \leq b_{I_1}, \; g_{\mathrm{nc}}(x_{\mathrm{nc}}) \leq 0\}$, where $A_{I_1}=[a_1^\top;\cdots;a_{n_1}^\top]$ and $b_{I_1} = [b_1; \cdots; b_{n_1}]$.
For FRS-funnel composability check, the objective of \autoref{eq:ddp} is set to $a(e)=g_{\mathrm{nc}}(x_{r\cdot \mathrm{nc}}(T) + e_{\mathrm{nc}}) \leq 0$ and $a(e) = a_i^\top (x_{r\cdot \mathrm{c}}(T) + \Pi(e)) \leq b_i$, for noncyclic and cyclic coordinates, respectively. The cyclic composability check should yield the translatable margin for the first funnel as $A_{I_1} \delta p \leq b_{\mathrm{FRS}}$. 
For each $\tau \in \{0, \cdots, T\}$, collision with the environment is also checked. For that, we construct a single safe flight corridor (SFC) in cyclic coordinate space around the local trajectory using \cite{sfc}. 
SFC is a large convex region built in the obstacle-free space, which we can use to bring a nonconvex trajectory optimization problem to a slightly more conservative but convex domain \cite{sfc0, sfc}.
Denote the obtained SFC by $\mathrm{SFC} = \{p \;|\; A_{\mathrm{SFC}}p \leq b_{\mathrm{SFC}}\}$ where $A_{\mathrm{SFC}}=[a_{\mathrm{SFC} \cdot 1}^\top; \cdots; a_{\mathrm{SFC} \cdot n_{\mathrm{SFC}}}^\top]$. Similarly, collision can be checked by letting $a(e)=a_{\mathrm{SFC} \cdot i}^\top \Pi(e)$.

\subsection{Funnel Loop Closure} \label{sec:adjustment}

Let the result from the previous step be denoted by an ordered set $(\mathcal{F}_1, \cdots, \mathcal{F}_{n_F})$, which is a valid funnel composition (but yet to be loop-closed). The objective of this step is to adjust the given funnels along the cyclic coordinates, so that $\mathcal{F}_{n_F}$ and $\mathcal{F}_1$ are connected, while not losing validity. 
We consider the situation in which $\mathcal{F}_i$ shall be translated by $\delta p_i \in \mathbb{R}^d$. Two constraints should be satisfied for each translation: collision avoidance, and funnel composability.

\begin{figure}
    \centering
    \includegraphics[width=0.8\linewidth]{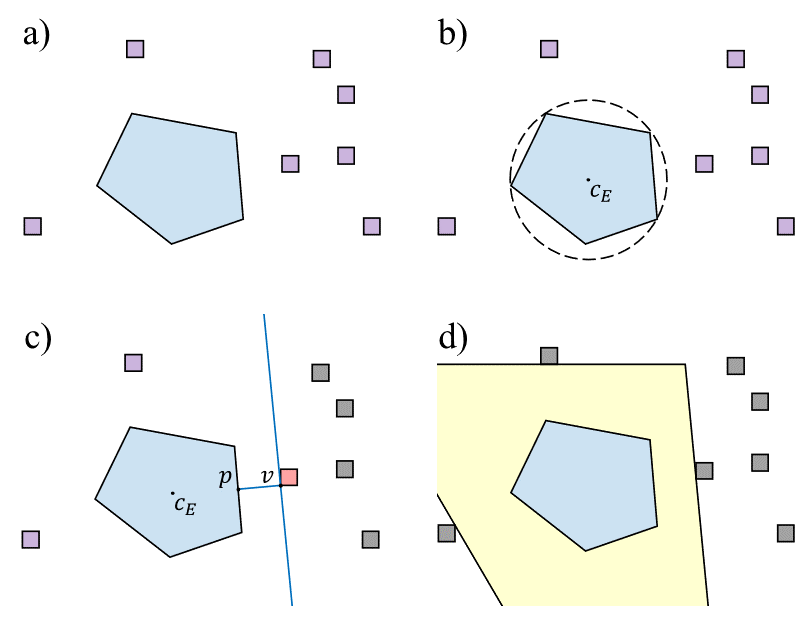}
    \caption{A graphical description of Algorithm \ref{alg:funneladjustment}. a) The target funnel encompassing shape $E$ is depicted blue. The purple squares represent the nearby occupied (or unknown) voxels in the open set. Voxels removed from the open set are colored gray. b) First, an encompassing circle (or sphere) of $E$ is found to obtain $c_E$. c) Starting from the nearest voxel (red), two points $p$ and $v$ are obtained to find a wall (depicted using blue line). Voxels outside the wall are removed from the open set. d) After finite iterations, open set is emptied, and the obtained walls form a convex region in which the funnel can freely translate (yellow polygon).}
    \vspace{-0.4cm}
    \label{fig:funneladjustment1}
\end{figure}

\subsubsection{Colision Avoidance Constraint}

For collision avoidance, we propose an algorithm that generates a convex \textit{adjustable area} of a funnel, in which the funnel can translate without collision.
We first represent each obstacle as a convex shape in $\mathbb{R}^d$ space, which is a voxel in most cases. The unknown space is also considered an obstacle, because the funnel loop must be built within regions known to be free.
Our algorithm (\autoref{alg:funneladjustment}) builds linear inequalities, one at a time, whose intersection constructs the adjustable margin. 
It initializes with the unadjusted encompassing polytope $E$ of a funnel, and the \texttt{open\_set} that contains neighboring obstacle yet to be considered. The \texttt{smallest\_sphere\_center} in \autoref{algline:smallest_sphere_center} returns the center $c_E$ of the smallest sphere that completely covers $E$. The loop in \autoref{algline:whilestart} iterates over the elements of \texttt{open\_set}, starting from the nearest obstacle from $c_E$, and finds a separating hyperplane between the funnel and the obstacle. Using the hyperplane, we find the linear constraint written $\{ \delta p \in \mathbb{R}^d \;|\; a^\top \delta p \leq b\}$ ($a \in \mathbb{R}^d$ and $b \in \mathbb{R}$), which guarantees $E$ to be collision-free when translated by $\delta p$. 
With an obstacle region $V$, we first find two points, each in $E$ and $V$, that are closest, by solving a QP problem with $2d$ decision variables:
\begin{equation} \label{eq:smallqp}
\begin{aligned}
    \underset{p, v \in \mathbb{R}^d}{\mathrm{min.}} & \quad \norm{p-v}_2^2 \\
    \mathrm{s. t.\;\;} & \quad p \in E, \quad v \in V.
\end{aligned}
\end{equation}
Since the two shapes are convex, we can conclude that $E$ can be translated in the direction of $v-p$ by distance $\norm{v-p}_2$ while not colliding with $V$, i.e.,
\begin{equation} \label{eq:cpd}
    \left(\frac{v-p}{\norm{v-p}_2}\right)^\top \delta p \;=\; a^\top \delta p \;\leq\; \norm{v-p}_2 \;=\; b
\end{equation}
provides a collision-avoiding bound for $\delta p$. 
We then remove obstacles that are farther in the $v-p$ direction than $V$ from the \texttt{open\_set}, as they are already not reachable by $E$ when \autoref{eq:cpd} is satisfied.
The obtained $a^\top$ and $b$ are added to the rows of $A_A \in \mathbb{R}^{n_A\times d}$ and $b_A \in \mathbb{R}^{n_A}$ 
This process iterates until the \texttt{open\_set} is emptied. 
A brief graphical explanation is provided in \autoref{fig:funneladjustment1}. 

\begin{algorithm}[t]
\caption{Funnel adjustable area computation}
\begin{algorithmic}[1] \label{alg:funneladjustment}
    \renewcommand{\algorithmicrequire}{\textbf{Input:}}
    \renewcommand{\algorithmicensure}{\textbf{Output:}}
    \REQUIRE Funnel bounding polytope $E$, \\
        \quad \; the set of nearby occupied voxels $\{V_1, \cdots, V_N\}$
    \ENSURE  $A_A$, $b_A$
    \STATE \texttt{open\_set} $\leftarrow \{V_1, \cdots, V_N\}$
    \STATE $c_E \leftarrow$ \texttt{smallest\_sphere\_center}($E$) \label{algline:smallest_sphere_center}
    \WHILE {\texttt{open\_set} is not empty} \label{algline:whilestart}
        \STATE $V \leftarrow$ element of \texttt{open\_set} closest to $c_E$
        \STATE $(p, v) \leftarrow \underset{p \in E, v \in V}{\mathrm{argmin}} \norm{v-p}_2$
        \STATE $a \leftarrow {(v-p)}/{\norm{v-p}_2}$, $b \leftarrow \norm{v-p}_2$
        \STATE Remove $V$ from \texttt{open\_set}.
        \FORALL {$V' \in$ \texttt{open\_set}}
            \IF {$a^\top (v'-v) > 0\;\;$ $\forall v' \in V'$}
                \STATE Remove $V'$ from \texttt{open\_set}.
            \ENDIF
        \ENDFOR
        \STATE $A_A \leftarrow [A_A ;\; a^\top]$, $b_A \leftarrow [b_A ;\; b]$
    \ENDWHILE 
    \label{algline:whileend}
\end{algorithmic}
\end{algorithm}

\subsubsection{Cyclic Composability Constraint}

For cyclic composability, \autoref{eq:cycliccondition_c} is directly used. Let $\mathcal{F}_i \rhd \mathcal{F}_j$. A point $p$ lies within the entrance of $\mathcal{F}_j$ translated by $\delta p_j$ if
\begin{equation}
    A_{I_j} (p - p_{I_j} - \delta p_j) \leq b_{I_j}.
\end{equation}
This should be satisfied by the exit of $\mathcal{F}_i$ translated by $\delta p_i$, with margin $r_{X_i}$.
\begin{equation}
    A_{I_j} (p_{X_i} - p_{I_j} + \delta p_i - \delta p_j) \leq b_{I_j} - r_{X_i}
\end{equation}
Additionally, the first funnel entrance should cover the FRS of the local trajectory:
\begin{equation}
    A_{I_1} \left(p_{I_1} + \delta p_1 - \Pi(x(T))\right) \leq b_{\mathrm{FRS}}
\end{equation}

Subject to the mentioned constraints, the following QP problem is solved to find feasible adjustment while minimizing jolty motions when switching between funnels:
\begin{equation} \label{eq:funneladj}
\begin{aligned}
    \mathrm{min.} & \quad \sum_{(i, j) \in \mathbb{I}} w_{i, j} \norm{(p_{X_i} + \delta p_i) - (p_{I_j} + \delta p_j)}_2^2 \\
    & \quad \quad + w_0 \norm{\Pi(x(T)) - (p_{I_1} + \delta p_1)}_2^2 \\
    \mathrm{s.t.} & \quad A_A^k \delta p_k \leq b_A^k \quad \forall k \in \{1, \cdots, n_F\} \\
    & \quad A_{I_j} (p_{X_i} - p_{I_j} + \delta p_i - \delta p_j) \leq b_{I_j} - r_{X_i} \quad \forall (i, j) \in \mathbb{I} \\
    & \quad A_{I_1} (p_{I_1} + \delta p_1 - \Pi(x(T))) \leq b_{\mathrm{FRS}}
\end{aligned}
\end{equation}
where $A_A^k$ is the adjustable area for the $k$-th funnel and $w_{i, j},\;w_0 \geq 0$ are nonnegative weights. The optimization problem \autoref{eq:funneladj} minimizes the magnitude of \textit{jump} required in switching between funnels in the cyclic coordinates. A good way to select the weights is to more heavily penalize jumps with lower indices, since they are more likely to be actually traversed than the ones with bigger indices.

\section{Experiment Results}

\begin{figure}[t]
    \centering
    \includegraphics[width=4.2cm]{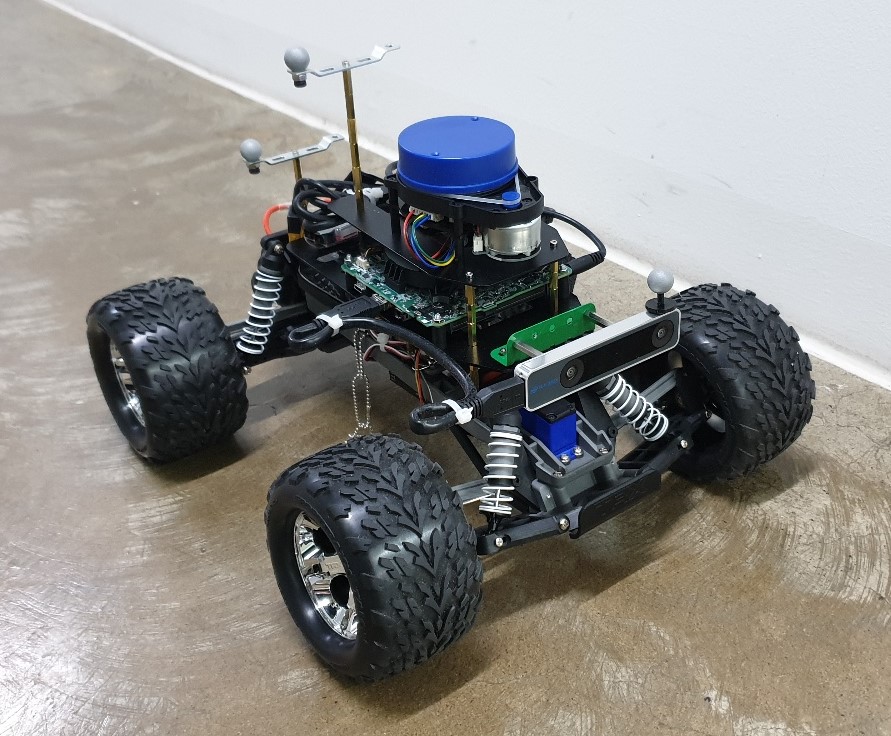}
    \caption{The ground rover used in the experiment.}
    \vspace{-0.4cm}
    \label{fig:car}
\end{figure}

To validate the proposed planning algorithm, an experiment was conducted using a ground rover shown in \autoref{fig:car}. The rover is an Ackerman steering robot equipped with an onboard computer (Intel NUC), a planar LiDAR sensor (YDLIDAR X4), and an IMU. The onboard computer has a 6-core CPU with base clock frequency of 1.10 GHz and 16 GB memory. The LiDAR scans the environment at 10 Hz.
The planning algorithm is implemented in C++, and OSQP \cite{osqp} is used for solving QPs. For real-time mapping and localization, Cartographer \cite{cartographer} is used. To provide a good initial guess to the SLAM module, pose estimates from the OptiTrack motion capture system and Intel Realsense Tracking Camera T265 are used. The map is updated every second.

\begin{figure}[t]
    \centering
    \includegraphics[width=0.4\textwidth]{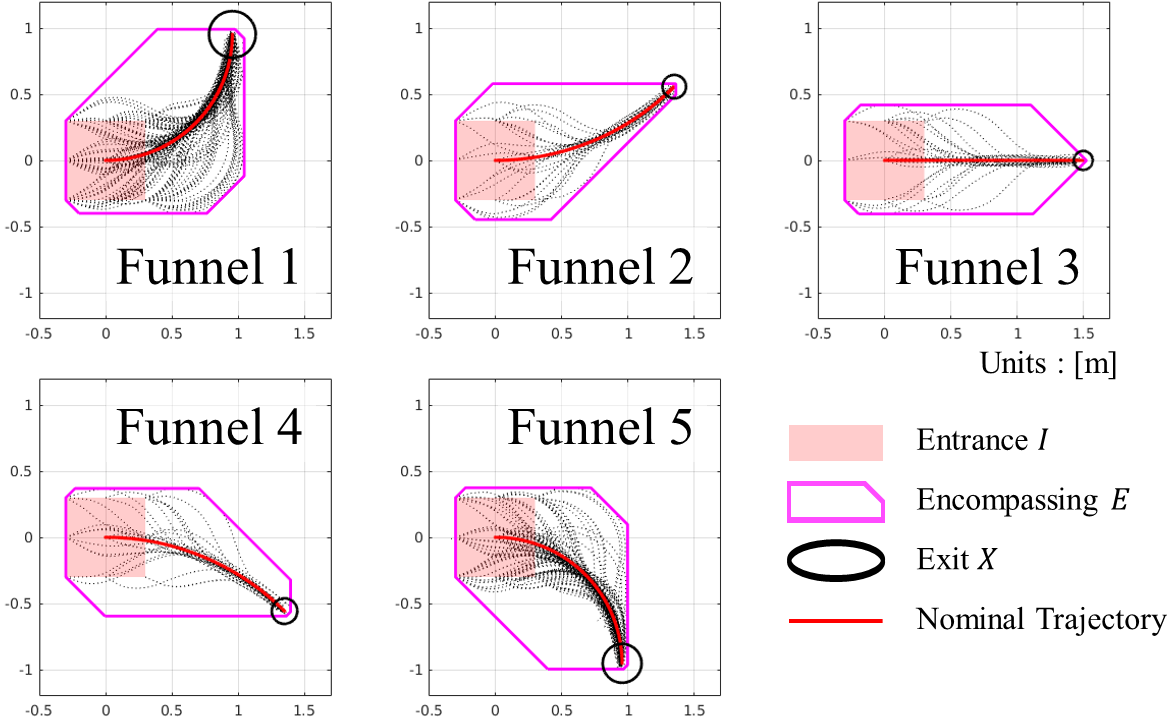}
    \caption{The first five funnels from the funnel library used in the experiment are depicted. The funnel library consists of 80 funnels in total, where the remaining 75 of them are the rotated ones the five funnels depicted in this figure. Dotted black lines represent the actual trajectory recorded from the funnel-building experiment.}
    \vspace{-0.4cm}
    \label{fig:funnels}
\end{figure}

\begin{figure*}
    \centering
    \begin{subfigure}{0.24\textwidth}
        \centering
        \includegraphics[width=\textwidth]{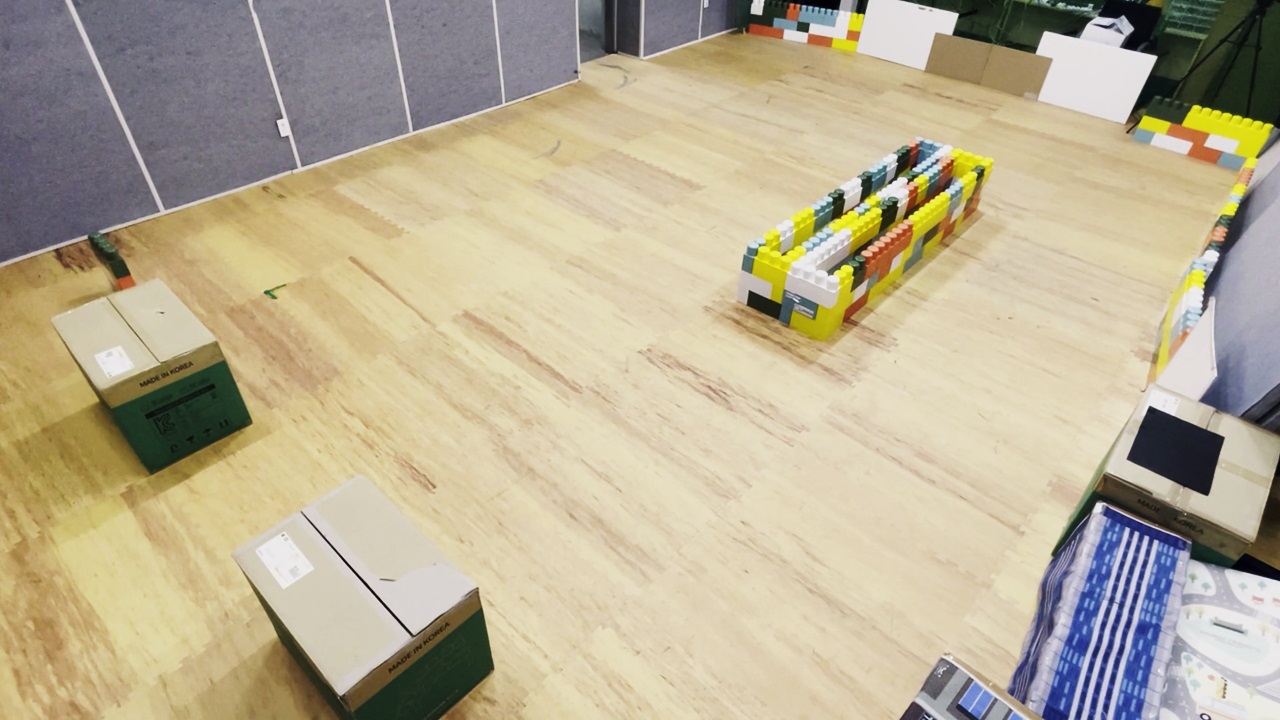}
        \caption{Scenario 1}
    \end{subfigure}
    \begin{subfigure}{0.24\textwidth}
        \centering
        \includegraphics[width=\textwidth]{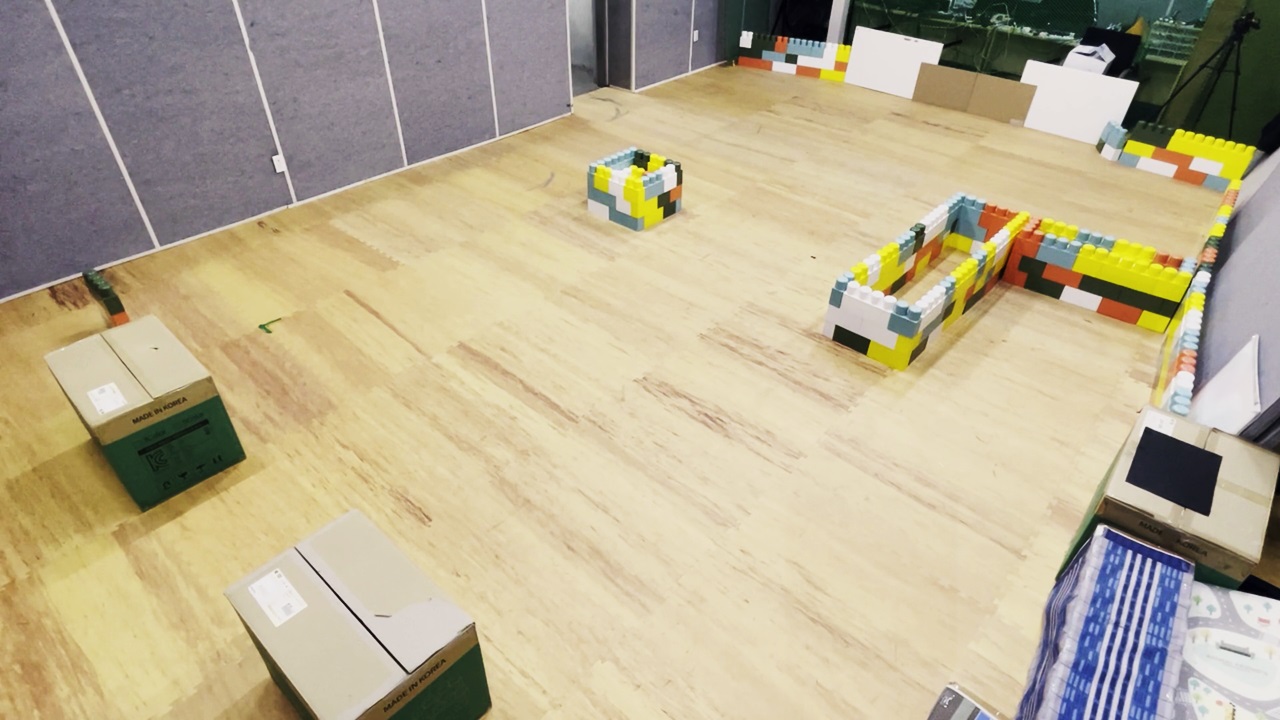}
        \caption{Scenario 2}
    \end{subfigure}
    \begin{subfigure}{0.49\textwidth}
        \centering
        \includegraphics[width=0.49\textwidth]{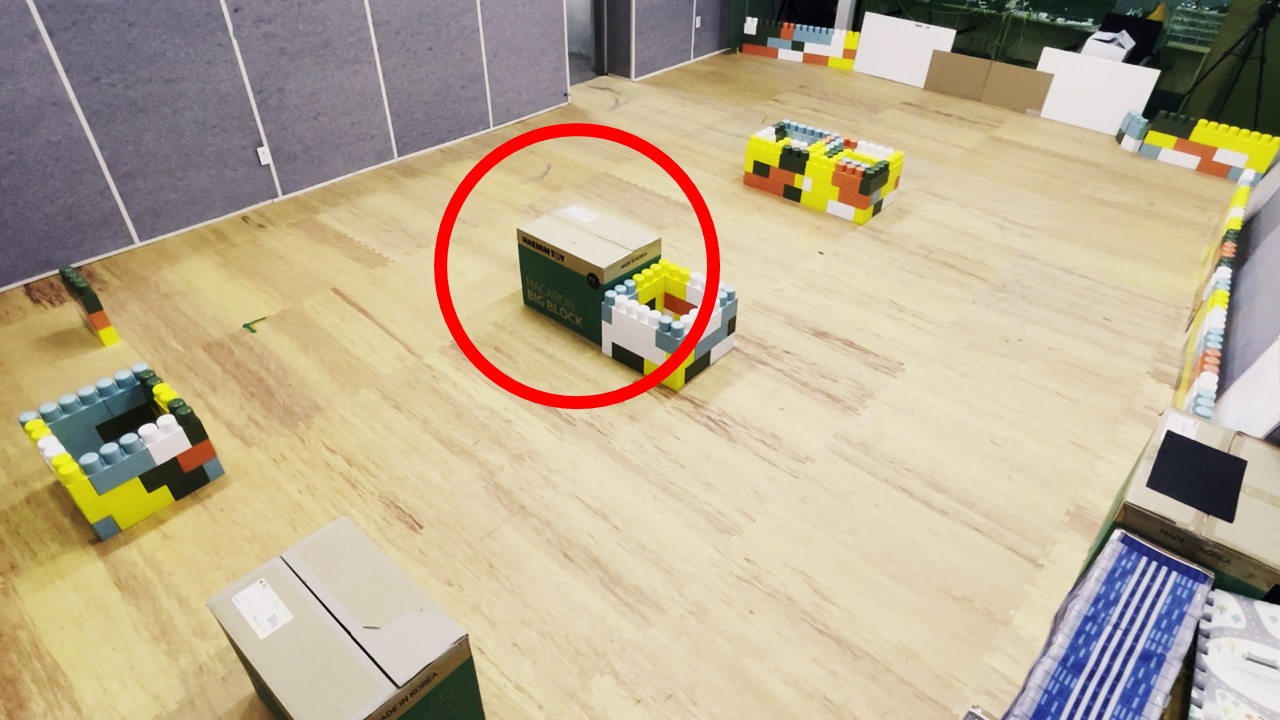}
        \includegraphics[width=0.49\textwidth]{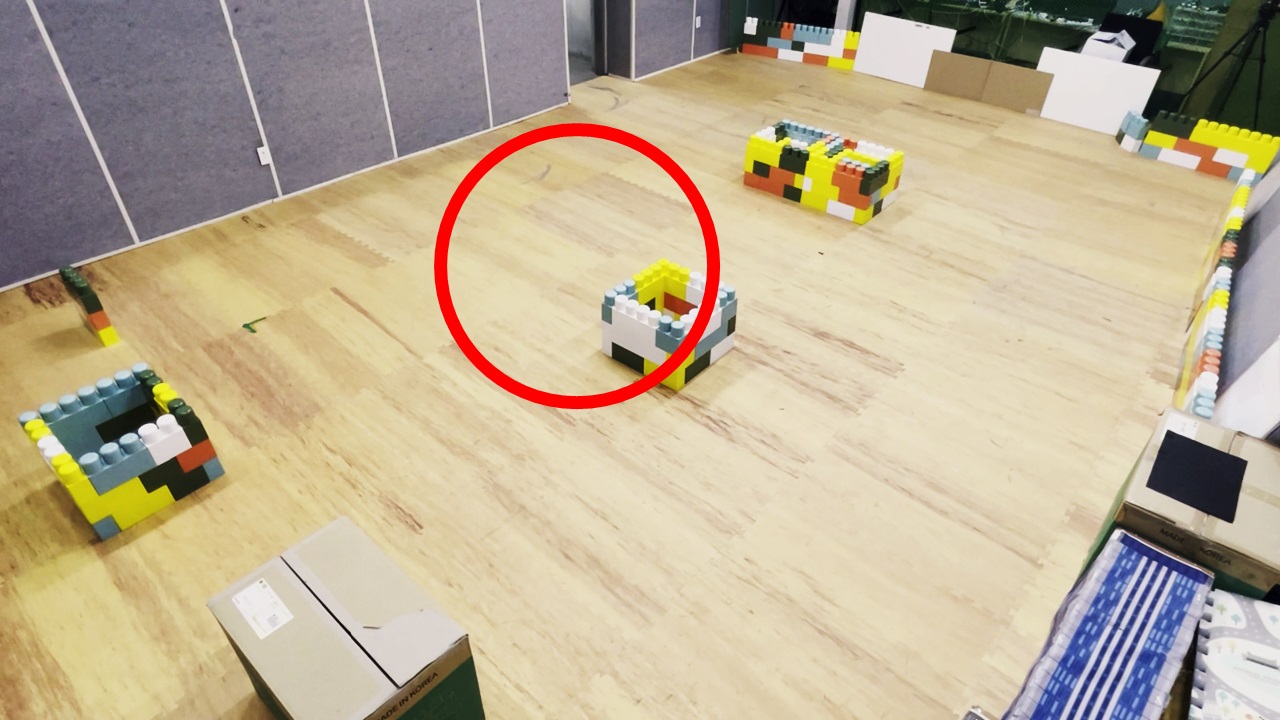}
        \caption{Scenario 3}
    \end{subfigure}
    \caption{Three scenarios tested in the experiment. (a) The first scenario consists of a corridor with both ends open. (b) The second also has a corridor, whose one side is however blocked. (c) The last environment consists of boxes sized about a meter in width and depth. One of them (red-circled box) is manually placed and removed repeatedly during the experiment.}
    \vspace{-0.2cm}
    \label{fig:scenarios}
\end{figure*}

\begin{figure}
    \centering
    \includegraphics[width=0.48\textwidth]{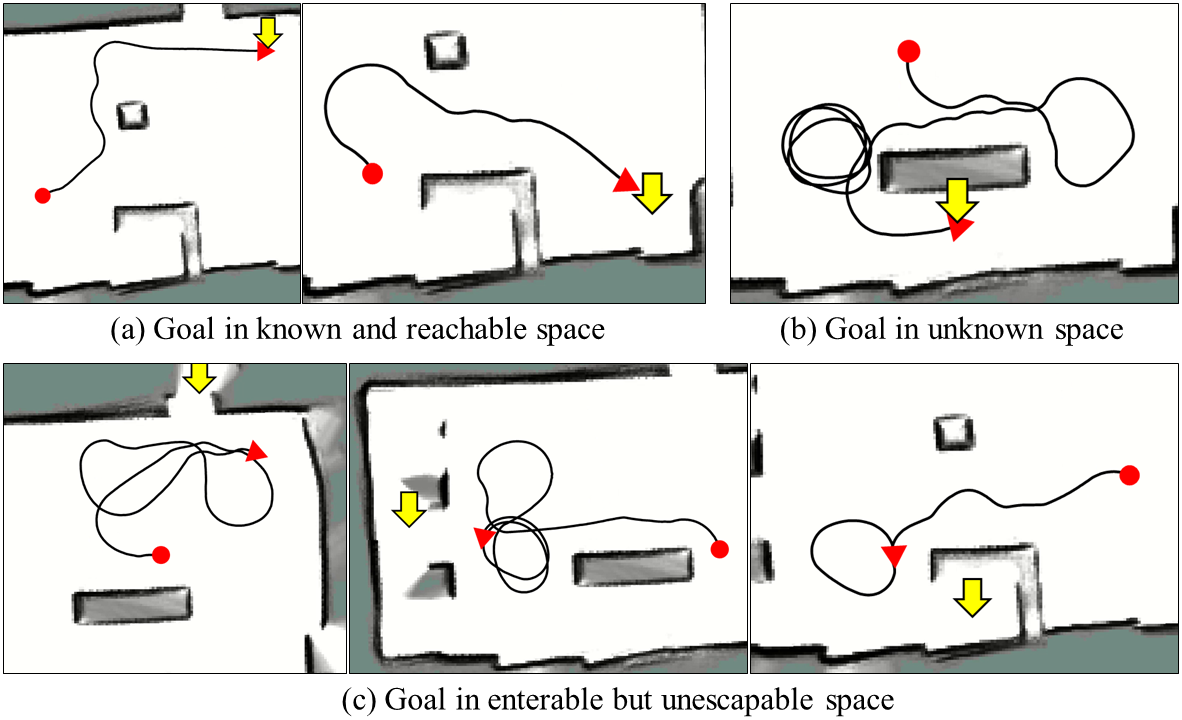}
    \caption{The trajectory snippets recorded during the experiment. The goal positions are marked using yellow arrows. Each trajectory (black curve) starts from a red circle ($\bullet$) and ends at an arrow head ($\blacktriangleright$).}
    \vspace{-0.4cm}
    \label{fig:trajectories}
\end{figure}

\begin{table}[b]
\renewcommand{\arraystretch}{1.3}
    \vspace{-0.4cm}
\caption{The values of the parameters used in the experiment}
\label{tab:parameters}
\centering
\begin{tabular}{c | c}
\hline
\bfseries Parameter & \bfseries Value\\
\hline\hline
$\delta t$ & $0.01$ s \\
\hline
$K_p$ & $[\;1.2, \;0, \;0; \;0, \;6.5, \;0.4 \;]$ \\
\hline
$K_d$ & $[\;0.15, \;0, \;0; \;0, \;3.0, \;0 \;]$ \\
\hline
$\underline{u}$ & $[\;0 \; \mathrm{m/s}; \;-1.1 \; \mathrm{m}^{-1}\;]$ \\
\hline
$\overline{u}$ & $[\;1.0 \; \mathrm{m/s}; \;1.1 \; \mathrm{m}^{-1}\;]$ \\
\hline
\end{tabular}
\end{table}

The kinematics of the ground rover system is governed by the following bicycle-model equation:
\begin{equation}
\begin{aligned}
    p_x(t+1) &= p_x(t) + v(t) \cos{\theta(t)} \cdot \delta t + w_1(t) \\
    p_y(t+1) &= p_y(t) + v(t) \sin{\theta(t)} \cdot \delta t + w_2(t) \\
    \theta (t+1) &= \theta (t) + v(t) \kappa(t) \cdot \delta t + w_3(t),
\end{aligned}
\end{equation}
where $x(t)=[p_x(t); p_y(t); \theta(t)] \in \mathbb{R}^3$, $u(t)=[v(t);\kappa(t)] \in \mathbb{R}^2$, and $w(t)=[w_1(t); w_2(t); w_3(t)] \in \mathbb{R}^3$ are the state, input, and disturbance vectors, respectively. The time discretization step $\delta t$ is a positive constant. The cyclic and noncyclic coordinates are $x_{\mathrm{c}}(t) = [p_x(t); p_y(t)] \in \mathbb{R}^2$ and $x_{\mathrm{nc}}(t) = [\theta(t)] \in \mathbb{R}^1$, respectively. We use the feedback controller in the following form to find the set point in speed $v_s(t)$ and turning rate $\kappa_s (t)$:
\begin{equation}
    \begin{bmatrix}
        v_s (t) \\
        \kappa_s (t)
    \end{bmatrix} = \mathrm{sat} \left(
        u_r(t) - K_p \cdot \begin{bmatrix}
            \Delta f \\ \Delta l \\ \Delta \theta
        \end{bmatrix} - K_d \cdot \frac{\mathrm{d}}{\mathrm{d}t} \begin{bmatrix}
            \Delta f \\ \Delta l \\ \Delta \theta
        \end{bmatrix}
    \right),
\end{equation}
where $K_p$ and $K_d$ are gain matrices in appropriate sizes, $\Delta f$ and $\Delta l$ represent the forward and leftward displacements of the robot measured from the reference trajectory, respectively. $\Delta \theta$ is the deviation of the heading angle $\theta$ with respect to the reference. The function $\mathrm{sat}:\mathbb{R}^2 \rightarrow \mathbb{R}^2$ clips the magnitude of the control commands to $\underline{u} \leq u(t) \leq \overline{u}$. The set points are forwarded to the low-level controller, which converts the set points to motor commands. The actual values of the parameters are summarized in \autoref{tab:parameters}.

To make the experiment more challenging, we restrict the speed of the reference trajectory to be fixed at 0.5 $\mathrm{m/s}$, and the turning rate (curvature) is limited to be less than 1.1 $\mathrm{m}^{-1}$. We however allow small jumps in the reference trajectory, and the tracking error due to jumps or external force is considered disturbance.
The funnels and disturbance bound used in the experiment are obtained through a preliminary experiment, in which the tracking error is measured while successively running randomly chosen nominal trajectories. \autoref{fig:funnels} shows the first five funnels used in the experiment, projected onto the cyclic coordinates.
For the global planner, jump point search (JPS) \cite{jps} is used. The local planner is sampling-based: we keep a trajectory library consisting of 7 local trajectories and select on that best tracks the global trajectory. The A$^\star$ algorithm is employed for the funnel loop candidate searching step, as it was one of the fastest among competing algorithms when used in the authors' implementation.

The experiment consists of three scenarios. In each scenario, the robot runs in different environments, which are shown in \autoref{fig:scenarios}. The goal positions change over time, and are given manually. In the experiment, the robot avoids the obstacle despite large tracking error, whose positions are \textit{a priori} unknown, while keeping its (tracking) reference speed at 0.5 $\mathrm{m/s}$. The following subsections provide discussions about some important snippets taken from the experiment results.

\subsubsection{Goal in Known and Reachable Space}

If the goal is given in known and reachable space, the planner acts like a normal receding horizon planner, because the local trajectory gets replaced by a new local trajectory before the robot enters the funnel loop.
\autoref{fig:trajectories}-(a) shows how the robot reaches the goal in environments where the funnel loops could be found easily.

\subsubsection{Goal in Unknown but Reachable Space}

\begin{figure}
    \centering
    \includegraphics[width=0.47\textwidth]{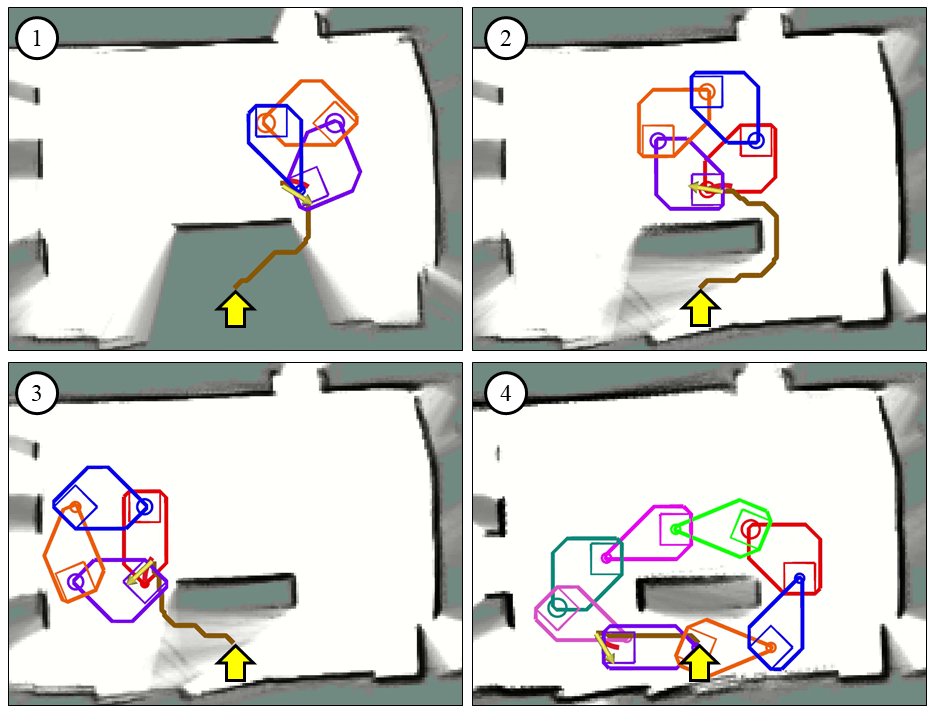}
    \caption{Funnel loops generated while trying to reach the goal in unknown space, behind the wall. The robot's poses are marked using thin yellow arrows. Brown lines represent the global trajectories that lead to the goal position at the big yellow arrow. Each funnel loops start with a local trajectory marked using a red curve.}
    \vspace{-0.4cm}
    \label{fig:funnelloops}
\end{figure}

\autoref{fig:trajectories}-(b) shows the trajectory of the robot when it is told to reach the goal in unknown space behind a wall. The robot cannot reach the goal at once, but while staying inside the funnel loop, it naturally explores the unknown space to build a funnel loop around the wall (\autoref{fig:funnelloops}).

\subsubsection{Goal in Enterable but Not Escapable Space}

Goals marked in \autoref{fig:trajectories}-(c) are reachable in the myopic sense but not potentially safe, because there is not enough space for the robot to turn around and escape. While funnel loops cannot be found in such cases, the robot does not reach the goal.

\begin{figure}[b]
    \centering
    \vspace{-0.4cm}
    \includegraphics[width=0.7\linewidth]{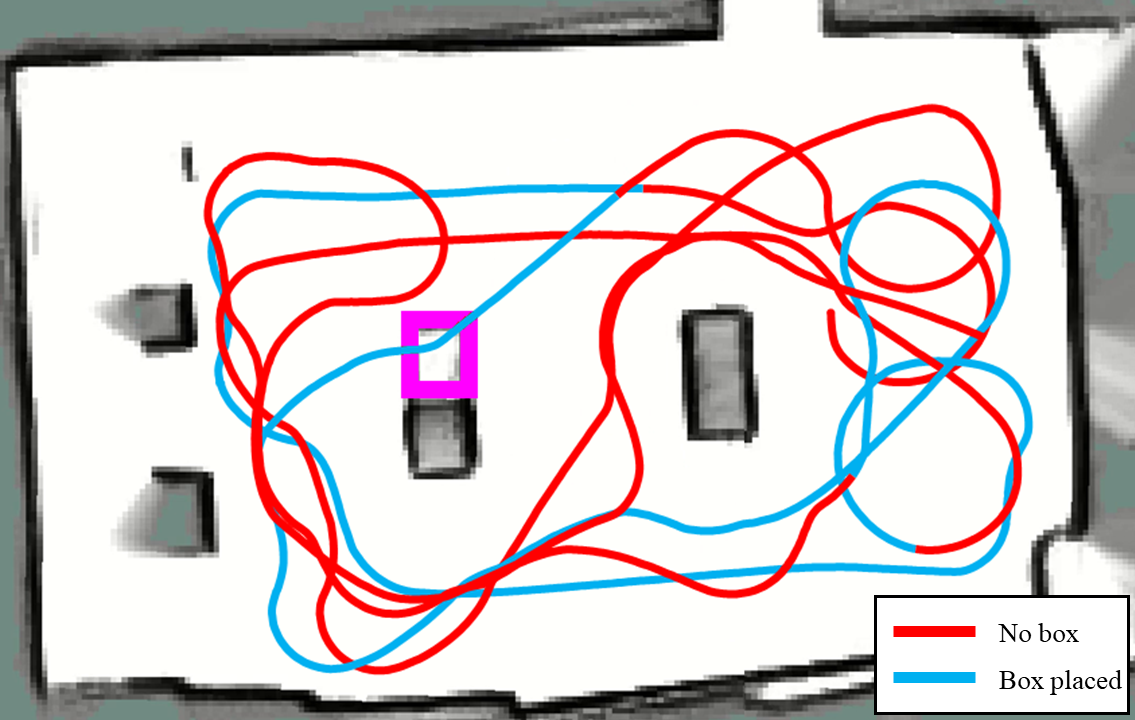}
    \caption{Trajectory data recorded during the third scenario. The position of the box is shown as the magenta rectangle. Red curves are the trajectory with the box in place, while blue curves are run while the box was removed.}
    \label{fig:boxtrajectory}
\end{figure}

\subsubsection{Nonstatic Environment}

The third scenario is designed to verify the resilience of the proposed algorithm to small map changes. The box shown in \autoref{fig:scenarios}-(c) is placed and removed repeatedly during the experiment. \autoref{fig:boxtrajectory} shows the recorded trajectory during a 3-minute run, during which the box was placed and removed twice. The robot succeeded to generate safe trajectories despite map changes, and also utilized the freed area during the box was removed.

\subsection{Computation Time Analysis}

For the sake of vehicle stability, the planner runs at 5 Hz in the experiment. However, a complete planning procedure finished in 60 ms (over 16 Hz) in average: 3 ms for global path planning, less than a millisecond for local planning, 47 ms for FRS calculation, 3 ms for funnel loop candidate searching, and 6 ms for the funnel loop adjustment step.

\vspace{-0.1cm}

\subsection{Remarks on Implementation}

The A$^\star$ algorithm is known to become faster with an optimistic heuristic, if optimality can be sacrificed. 
This is precisely the case, since the funnel loop is only used when the trajectory planning fails to find the next trajectory update and hence there is very little need to make the cyclic funnel sequence short. Thus, we used the heuristic which is ten times the consistent heuristic, which reduced the number of searches roughly in half.

Although the proposed algorithm can be run in real-time, the computation time is non-negligible. Thus, the local trajectory should start at the future state expected after the computation time estimation. 

\vspace{-0.1cm}
\section{Conclusion}

In this paper, we presented a planning algorithm that can be run under disturbances in unknown environments, while guaranteeing safety without emergency brakes. We first start by planning global and local trajectories. 
A loop of sequentially composed funnels is constructed starting from the end of the local trajectory. Requiring the FRS of the local trajectory to lie within the funnel loop entrance guarantees that the system can be driven robustly into the funnel loop, in which it can stay permanently without collision, seeking for the chance of trajectory update.
Experiment results showed that the planner can generate safe trajectories in real-time on onboard computers. It also demonstrated that the proposed algorithm is resilient to map changes, as the funnel loop is built locally.

Future work may include: enhancing the global planner so that  it does not repeat planning through unreachable spaces; exploiting other symmetries in the funnel planning step, e.g., rotation; and extending this algorithm to multi-agent systems.

\addtolength{\textheight}{-12cm}   








\bibliography{IEEEabrv, references}

\end{document}